# ECG Beats Classification via Online Sparse Dictionary and Time Pyramid Matching

Nanyu Li[1], Yujuan Si[1,2,*]
[1]Zhuhai College of Jilin University
Zhuhai, China
e-mail:13631268297@163.com
e-mail: siyj@jlu.edu.cn

Duo Deng[2], Chunyu Yuan[2]
[2]College of Communication Engineering
Jilin University
Changchun, China
e-mail: 1259247976@qq.com, 1572207574@qq.com

*Abstract*—Recently, the Bag-Of-Word (BOW) algorithm provides efficient features and promotes the accuracy of the ECG classification system. However, BOW algorithm has two shortcomings: (1). it has large quantization errors and poor reconstruction performance; (2). it loses heart beat's time information, and may provide confusing features for different kinds of heart beats. Furthermore, ECG classification system can be used for long time monitoring and analysis of cardiovascular patients, while a huge amount of data will be produced, so we urgently need an efficient compression algorithm. In view of the above problems, we use the wavelet feature to construct the sparse dictionary, which lower the quantization error to a minimum. In order to reduce the complexity of our algorithm and adapt to large-scale heart beats operation, we combine the Online Dictionary Learning with Feature-sign algorithm to update the dictionary and coefficients. Coefficients matrix is used to represent ECG beats, which greatly reduces the memory consumption, and solve the problem of quantitative error simultaneously. Finally, we construct the pyramid to match coefficients of each ECG beat. Thus, we obtain the features that contain the beat time information by time stochastic pooling. It is efficient to solve the problem of losing time information. The experimental results show that: on the one hand, the proposed algorithm has advantages of high reconstruction performance for BOW, this storage method is high fidelity and low memory consumption; on the other hand, our algorithm yields highest accuracy in ECG beats classification; so this method is more suitable for large-scale heart beats data storage and classification.

## I. INTRODUCTION

ECG Electrocardiogram (ECG) is one of the most common tools for the diagnosis of cardiovascular diseases. ECG consists of a series of heart beats (from the beginning of the P wave to the end of the T wave). The classified storage of the heart beats is an important basis for the diagnosis of cardiovascular diseases. The traditional manual classification is time-consuming and laborious; ECG classification system based on machine learning can play an important role in long-time monitoring [1] of patients with cardiovascular disease. The high accuracy of abnormal beats detection is extremely important for patients with cardiovascular diseases, at the same time, due to the large amount of data generated by long-time monitoring; an efficient beats compression algorithm is also urgently needed. Based on this motivation, we performed the related algorithm research and design an algorithm model which can efficiently store all the heart beats and detect various abnormal heart beats sensitively.

Nowadays, The feature extraction of fixed points on the time domain [2], frequency domain[3], and morphology [4] have certain limitations, sometimes the heartbeats may not contain some waves, and lead to the failure of extracting the corresponding feature points, and the robustness in the algorithms is not strong and susceptible to noise. The Bag-Of-Word (BOW) [5] can solve this problem by using statistical dictionaries to learn features from data sets; it also achieves ideal performance in heart beats classification [6], but the BOW model has two disadvantages: one is losing the time information of the beat, and the other is the poor reconstruction performance. For ECG systems, the BOW feature loss beat time information will lead to confusion between abnormal beats and normal beats in classification, thereby reducing the recognition rate of abnormal beats. BOW encoding has low fidelity, and that is the reason why it cannot be an efficient compression storage method used in the ECG system, therefore we must build another storage algorithm [7] to solve this problem, so if we can use one algorithm to solve the above two problems at the same time, it can effectively reduce the time computational complexity of ECG system. Fortunately, Kai Yu et al. [8] have proved that sparse coding has better reconstruction performance than BOW coding, and is more separable. In this paper, we propose online learning method to construct the sparse dictionary, use Online Dictionary Learning [9] to update the dictionary, and compute the coefficients with Feature-sign method [10], which can accelerate the speed of convergence, then we store each beat as the sparse coefficients matrix. On the other hand, Ponce J et al. have proved that spatial pyramid matching (SPM) [11] can prevent loss of spatial information in computer vision field. Based on this enlightenment, we construct Time Pyramid to calculate the dictionary coefficients and obtain more efficient feature representation by stochastic pooling [12]. Experiments show that the ECG coding stored by our method has high fidelity and low memory consumption, the average of reconstruction error is only 0.18%, with average compression ratio of 55.5%. In heart beat classification, the accuracy of sick heart beats is more than 90%, and the accuracy of Pacing heart beats is nearly 100%. In Section II, the disadvantages of the BOW method are described in detail. In Section III, corresponding the disadvantages of the BOW, we propose a solution by online sparse dictionary learning and Time Pyramid Matching. In Section IV, we discuss proposed algorithm from five modules: pretreatment, sparse dictionary, online learning, ECG beats compression, and Sick heart beats recognition. In Section V, we show the experimental procedures and experimental results. Finally, in Section VI, we summarize the main results and look forward to the paper.

## II. THE DISADVANTAGES OF BOW MODEL

BOW model is an important method developed in the text field [13], which is widely used in the fields of computer vision [14] and signal processing [5]. It uses the local features of the signal and image sets to construct the dictionary; dictionary and coefficients optimize by K mean, then sum up the coefficients, each signal or image is displayed by histogram. Histogram is an effective feature. So in ECG system, the formula of the BOW model is represented as:

$$\min_{D \in R^{d \times k}, \; X \in R^{k \times n}} \sum_{i=1}^{n} \|y_i - Dx_i\|_2 \quad (1)$$

$$\text{S. t. card }(x_i) = 1, \; |x_i| = 1, \forall i, x_i \geq 1,$$

$Y = [y_1, y_2 \ldots y_n] \in R^{d \times n}$ are local wavelet features of the ECG beats. Then D is the dictionary, $X = [x_1, x_2 \ldots x_n] \in R^{k \times n}$ are the coefficients, each vector $x_i$ has only one element of 1, and the others are 0, which ensures that each signal feature corresponds to only one dictionary atom(column of the dictionary). Sum up $x_i$, and each beat is represented by histogram, then classified by support vector machine (SVM) [15], but there are some problems of BOW in ECG system.

• BOW model uses the K mean algorithm, which forced constraints card $(x_i) = 1$, ignores the relationships between different dictionary atoms, and exists large quantization errors. In ECG system, reconstruction error of each ECG beat reaches 17%, as shown in Figure 1 (a), so the encoding fidelity is low and cannot be used for beats compression.

• The histogram ignores the time information of beats, which can lead to a situation like Figure 2 (a), In Figure 2(a): the left upper picture shows two abnormal beats (red and green lines) and a normal beat (blue line), the upper right picture shows the abnormal beat histogram (red bar), the lower left picture shows the normal histogram (blue bar), and the lower right picture shows the abnormal histogram (green bar). Although the abnormal heart can be distinguished easily by human eyes (the left upper picture), but the histogram of abnormal heart beat (red bar) is more similar to the histogram of normal beats (blue bar) rather than the histogram of abnormal heart beat (green bar), which will lead to misjudgment of classifier, the reason is that the normal heart beat and abnormal heart beat still exists local similarity in the wavelet domain.

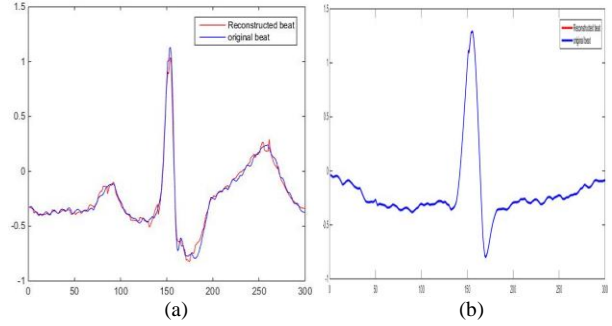
Figure 1. ECG reconstruction performance of Bow model and sparse dictionary model

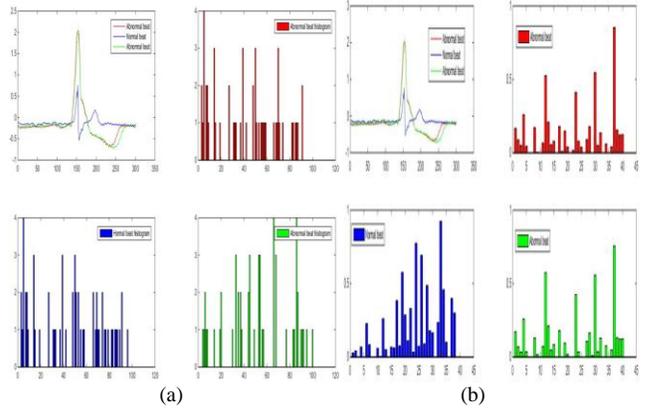
Figure 2. The feature of Bow model and sparse dictionary model

## III. ONLINE SPARSE CODING AND TIME PYRAMID MATCHING

To solve the above problems, we propose to use the online sparse dictionary learning to improve the traditional BOW heart beat model. Sparse dictionary as an improved model of BOW has been proved useful for image recognition [16] and signal clustering [17]. Sparse coefficient has better reconstruction performance, and avoids the large quantization errors, as shown in Figure 1(b), using this approach on heart beat system can hold the reconstruction error of the heart beat to a minimum. The sparse coefficients have a small number of non-zero elements. Each beat is stored by sparse coefficients, which reduces memory consumption. Furthermore, our algorithm based on online learning and $\ell_1$ norm optimization, is more suitable for large-scale heartbeat calculations. Spatial Pyramid Matching, as a method preventing histogram representation loses image spatial information. Inspired by this, we use the Time Pyramid Matching that is, constructing the pyramid in the time line for the sparse coefficients of each ECG beat. Thus, we obtain the features that contain the beat time information by time stochastic pooling. It is effective to solve the problem that the heartbeat signal loses time information on the BOW model., As shown in Fig. 2 (b), there are big differences between histogram representation of the abnormal heart beat (red and green) and the normal heart beat (blue), so we can distinguish the heart beat well through the classifier.

## IV. OUR APPROACH

### A. Pretreatment

Base on [18], we use bior2.6 wavelet decomposition to remove the baseline drift and noise, and detect QRS, P, T waves and their boundary, as shown in Figure 3:

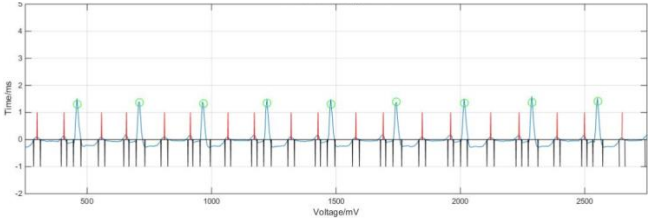

Figure 3. The wave detection in ECG signal

Where red line is for the P or T wave peak, green circle is for the QRS wave peak, and black lines indicate the boundaries of each wave. We split each heart beat (the start point of P wave to the end point of T wave) and unify [19] each beats for 300 sampling points, and then normalize them.

We use the sliding window to capture the local wavelet feature of each heart beat, and establish heart beat local features set Y=$[y_1, y_i \dots y_\eta]$ (the number of heart beat: Γ)Y∈ $R^{d\times\eta}$, and random select Y=$[y_1, y_2 \dots y_n]$, Y∈ $R^{d\times n}$, as the training set.

### B. Sparse Dictionary

We use the training set Y to construct dictionary, and the formula is expressed as follows:

$$\min_{D\in R^{d\times k},\ X\in R^{k\times n}} \sum_{i=1}^{n} \|y_i - Dx_i\|_2$$
$$S.t.\ \forall j = 1, \dots, k, \|d_j\|_2 = 1 \quad (2)$$
$$\forall i = 1, \dots, n, \|x_i\|_0 \leq T$$

$D \in R^{d\times k}$ (k > d) is unknown over complete dictionary, X=$[x_1, x_2 \dots x_n]$ are unknown spare coefficients, the number of nonzero elements in a vector $x_i$ is not greater than T, Eq.( 2) is an NP hard problem, and it is difficult to find an optimal solution. Fortunately, the current theoretical studies [20], [21] reveal that if the solution is sparse enough, solution of Eq. (2) equals to that of Eq. (3), which replaces $\ell_0$ norm with $\ell_1$ norm, i.e.:

$$\min_{D\in R^{d\times k},\ X\in R^{k\times n}} \sum_{i=1}^{n} \frac{1}{2}\|y_i - Dx_i\|_2^2 + \lambda\|x_i\|_1 \quad (3)$$
$$S.t.\ \forall j = 1, \dots, k, \|d_j\|_2 = 1$$

Although the equation has two unknowns, we should know how to efficiently construct its dictionary and how to compute its sparse coefficients. What is more, when X and D are fixed, this problem will become a convex optimization problem. The methods for compute coefficients includes: LARS [22] and feature-sign [10]. The methods for update dictionary includes: Online Dictionary Learning [9] and Lag-Dual [10].

### C. Online Learning

As shown in Algorithm 1, we use the Online Dictionary Learning to update the dictionary, the Feature-sign algorithm to compute the coefficients. Based on the acceleration principle of online learning, we divide the training set into several batches, such as $\dot{Y}$= $[y_1, y_2 \dots y_\xi]$ (mini batch size= ξ), use our approach (online dictionary learning and feature-sign) to optimize each batch, return the $\dot{D}$, $\dot{X}$, and the $\dot{D}$, $\dot{X}$ become the initial value of the next batch optimization. This principle can speed up convergence, and reduces memory consumption.

**Algorithm 1.** Online sparse dictionary learning
**For t=1 to T，mini batch size=ξ**
**Input：** $\dot{Y}$=$[y_1, y_2 \dots y_\xi]$~P(Y), $\dot{D}\to D$
1: $j = \arg\max_j \frac{\partial\|y_i - Dx_i\|_2^2}{\partial x_i^{(j)}}$ , $g^{(j)} = \frac{\partial\|y_i - Dx_i\|_2^2}{\partial x_i^{(j)}}$

2: if $g^{(j)} > \lambda$, then $x_i^{(j)} = \frac{\lambda - g^{(j)}}{D_j^T D_j}$;

else if $g^{(j)} < -\lambda$, then $x_i^{(j)} = \frac{-\lambda - g^{(j)}}{D_j^T D_j}$;

else if $|g^{(j)}| \leq \lambda$, then $x_i^{(j)} = 0$, break;
3: **while true，do**
4: $\delta = find(x_i \neq 0)$; $D_\delta = D(:,\delta)$;

$\frac{1}{2}\frac{\partial\|y_i^{(\delta)} - D_\delta x_i^{(\delta)}\|_2}{\partial x_i^{(\delta)}} + \lambda sign(x_i^{(\delta)}) = 0 \to$
$x_{new}=(D_\delta^T D_\delta)^{-1}(D_\delta^T y_i^{(\delta)} - \lambda sign(x_i^{(\delta)}))$
5: if $sign(x_i^{(\delta)}) == sign(x_{new})$, then $x_i^{(\delta)} \leftarrow x_{new}$, break
   else if perform a discrete line search;
6: Initialization $X_t \leftarrow \dot{X} = [x_1, x_2 \dots x_\xi], A_0 = \emptyset, B_0 = \emptyset$;
   $A_t = A_{t-1} + X_t X_t^T, B_t = B_{t-1} + Y_t X_t^T$;
   A=$[a_1, \dots, a_k]\epsilon R^{k\times k}$, B = $[b_1, \dots, b_k]\epsilon R^{d\times k}$

7: $D_t = \arg\min_{D\in R^{d\times k}} \frac{1}{t}\sum_i^t \frac{1}{2}\|Y_i - DX_i\|_2^2 + \lambda\|X_i\|_1$
$= \arg\min_{D\in R^{d\times k}} \frac{1}{T}(\frac{1}{2}Tr(D^T DA_t) - Tr(D^T B_t))$
8: **for j=1 to K do;**
   $d_j = \frac{1}{\max(\|u_j\|_2,1)}[\frac{1}{A[j,j]}(b_j - Da_j) + d_j]$;
9: **end for**
**Output:** $D_t \to \dot{D}, \dot{X}$
**End for**

In the Alg. 1, g is the gradient of $\|y_i - Dx_i\|_2^2$ in terms of $x_i$, $g^{(j)}$ is the maximal element in the vector g. $x_i^{(j)}$ is the $j^{th}$ element of the vector $x_i$, but δ is the vector, so $x_i^{(\delta)}$ is still a vector. For dictionary atomic update, the step size is $\frac{1}{A[j,j]}$, $(b_j - Da_j)$ is the gradient negative direction; $\frac{1}{A[j,j]}(b_j - Da_j) + d_j$ is the gradient descent method. Compared with other sparse dictionary methods: KSVD [23] SVD decomposition and Lag-Dual [10] Hessian matrix solution, the algorithm complexity is much lower.

*D. Heart Beat Compression*

The local wavelet features Y=[$y_1, y_i \ldots y_\eta$] (the number of heart beats: $\Gamma$ ) are compressed and stored X = [$x_1, x_2 \ldots x_\eta$] $\in R^{k \times \eta}$ by using the above dictionary learning, so the heart beats are stored as a k× $\Omega$ matrix, $\Omega = \frac{\eta}{\Gamma}$, sparse dictionary only stores the coordinate of non-zero elements, thereby reducing the memory usage and achieving compression.

$$x_{ij} = \begin{bmatrix} x_{1\,1} & 0 & \ldots & 0 \\ 0 & 0 & 0 & 0 \\ 0 & x_{3\,2} & \cdots & 0 \\ \vdots & \vdots & \vdots & \vdots \\ 0 & x_{k\,2} & 0 & x_{k\,\Omega} \end{bmatrix} \quad (4)$$

In addition, sparse matrix is easy to store, it is easy to correct even there is distortion in the transmission process; sparse matrix is also easy to restore each beat the computational complexity is small.

*E. Sick Heart Beats Recognition*

In order to solve the problem that the traditional histogram features represent lost time information, similar to SPM method, we take the absolute value X=|X| to prevent the different $x_i$ appears offset phenomenon during the statistics period, then we construct a multi-layer pyramid, we get $\Pi$ blocks from each heart beast, each blocks represents periods of different sizes. We count $x_i$( suppose the number of they is m) corresponding each block, and do stochastic pooling [12], and select more effective coefficient, finally, the coefficient of each block is summed, all the steps are as follows:

$$z = \sum_{j=1}^{\Pi} z_j; \; z_j = X_l \;; \; z \in R^k$$
$$\text{where } l \sim P, \; P = \frac{X_m}{\frac{1}{M}\sum_{m=1}^{M} X_m}, \; X_m \in R^{k \times m} \quad (5)$$

z is a histogram representation which improved by time pyramid and stochastic pooling. Compared to the traditional histogram, it is more sensitive to the time information of heart beats. Stochastic pooling also improves the robustness of the feature (histogram representation). We arranged the histogram representation of all the heart in matrix form Z=[$z_1, z_2 \ldots z_\Gamma$]$\in R^{k \times \Gamma}$, then randomly extract the data (i = 1 ... n) as a training data , use support vector machine (SVM) [15] for classification and then find the super-plane which can separate different types of heart beat, SVM formula is as follows:

$$\min_{w,b} \frac{1}{2}\|w\|^2 + C\sum_{i=1}^{m} \xi_i$$
$$\text{S. t. } y_i(w^T \Phi(z_i) + b) \geq 1 - \xi_i, \forall i \quad (6)$$
$$\xi_i \geq 0, \forall i$$

X is the defined hyper plane, b is its offset, w is its coefficient, $y_i$ is the label corresponding to each $z_i$, $\Phi$ is the mapping function, $\xi_i$ is the relaxation coefficient, C is the penalty parameter to ensure that there are minimized training errors between two different categories, the max geometric interval is $\frac{1}{\|w\|}$. We use the Lagrange dual solver to solve the Eq. (6), Based on the KKT [24] condition and the kernel function [25] method, the original equation is transformed into:

$$L_{svm} = \min_\alpha \frac{1}{2}\sum_{i=1}^{m}\sum_{j=1}^{m} y_i y_j \alpha_i \alpha_j K(z_i, z_j) - \sum_{i=1}^{m} \alpha_i$$
$$\text{S. t. } \sum_{i=1}^{m} \alpha_i y_i = 0 \quad (7)$$
$$0 \leq \alpha_i \leq C, \forall i$$

here we use the RBF kernel function [26] satisfying the Mercer theorem [25] :

$$K(a,b) = \exp(-\gamma\|a - b\|^2) = \phi(a)^T \phi(b) \quad (8)$$

Based on this, we only need PSO [26] method to automatically find the optimal parameters: C (Penalty coefficient) and Y(kernel function coefficient), then optimize the solution of Eq. (8), and then the desired hyper plane can be obtained. Figure 4 is a flow chart of the complete ECG beats storage and identification system in this paper. After a series of pretreatments for ECG signal, including noise removal, alignment; we extract its local wavelet feature, construct sparse dictionary, achieve compress storage; and represented by histogram after Time Pyramid Matching, then abnormal heart beat (red) can be identified by SVM.

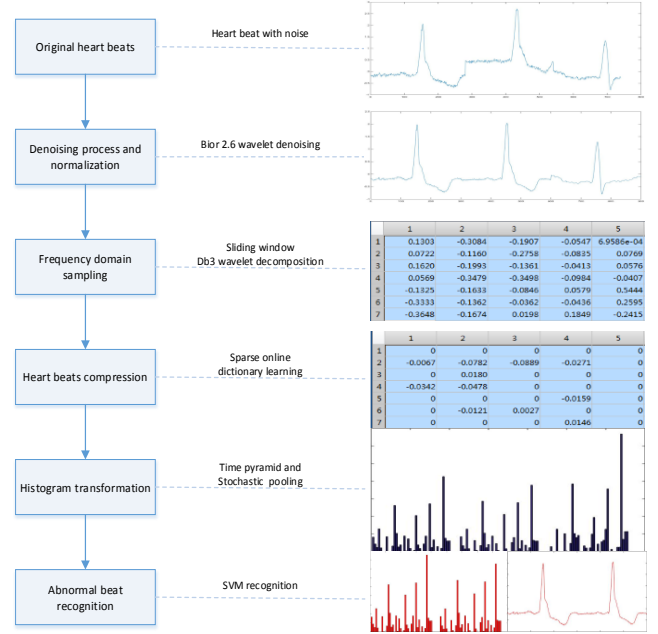

Figure 4. The flow chart of ECG heart beat system: compression storage and abnormal heart beat identification

## V. EXPERIMENTS AND RESULTS

### A. Database

The ECG beats data in our experiments are from MIT-BIH Physionet Database [27]. The database includes 48 persons' ECG data, half hour for each person. According to the pretreatment in the section.4, various types of heart beats are isolated from the Electrocardiogram. We select the most common 6 categories beats: Normal heart beat(N), Pacing heartbeat (/), Atrial premature beats (A), Ventricular premature beats (V), Right bundle branch block (R), Left bundle branch block (L) for heart beats reconstruction, compression storage and classification experiments.

### B. Experiment Setting

We randomly extract 1159 heart beats where each category is taken as a training set and rest as a test set. According to the description in Section IV, we get sparse matrix by online sparse dictionary learning, compared with the BOW model [5], we evaluate our method's reconstruction errors and compression ratio, then get the histogram representation by Time Pyramid matching. Finally use SVM classifier to do classification, compared with the BOW model, VQ model [6], the traditional method of extracting feature points (ECG morphology and temporal features) [28], we evaluate our method's beats classification accuracy. Finally we verify whether our method can solve BOW problems in ECG system and achieve the best performance at present.

### C. Experiment Result

•Heart beat compression

According to the linear properties of the dictionary learning and the wavelet transform, we reconstruct heart beats by coefficients, measure the difference between the reconstructed heart beat and the original heart beat, and evaluate the fidelity of the dictionary compression. The quantization error is defined as:

$$\text{Err}=\text{mean} (\sum \frac{\|M-N\|_2}{\|N\|_2}) \quad (9)$$

where M is the reconstructed signal and N is the original signal. The results of the experiment are shown in Table I. The reconstruction effect of BOW is shown in Figures 5-6, and the reconstruction effect of Propose method is shown in Figures 7-8, which evaluate whether our propose method can be used for ECG compression. We can see that our algorithm has better reconstruction performance than BOW, and the error is reduced to nearly 0.1%. The results reveal that the sparse dictionary solves the quantization error problem in ECG system, and can be used as the heart beats compression algorithm. We calculate the compression ratio of our sparse dictionary algorithm, which is defined as:

$$\text{Cr}=\text{mean}(\sum \frac{n-m}{n}) \quad (10)$$

We use the sparse dictionary to compress the wavelet features for each beat, Where m is the number of nonzero elements in sparse matrix each beats (we only stores the coordinate and value of non-zero elements), and n is the number of elements each beats (300 samples). The results are shown in Table II, which reveal that our method gets better compression performance. Compared to the direct storage of heart beats, sparse dictionary compression can reduce nearly half of the memory consumption. For large data ECG systems, it has far-reaching significance.

TABLE I. ECG BEATS RECONSTRUCTION ERROR

|  | N | / | A | V | R | L | Mean |
|---|---|---|---|---|---|---|---|
| Our method | 0.12% | 0.19% | 0.21% | 0.17% | 0.21% | 0.18% | 0.18% |
| BOW | 18% | 10% | 22.3% | 17.4% | 15% | 15.8% | 16.42% |

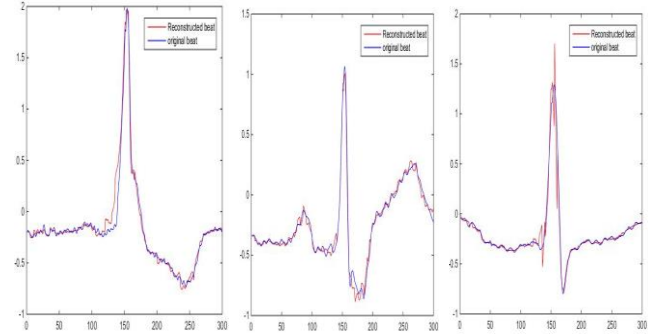

Figure 5. In the figure, from left to right respectively are reconstruction effects of atrial premature beats (A), left bundle branch block (L), normal heart beat (N)

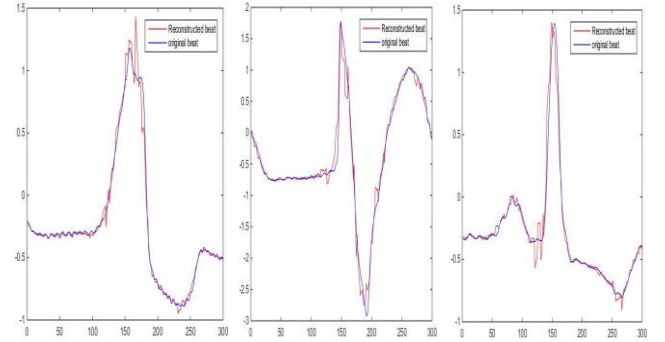

Figure 6. From the left to the right respectively are: the reconstruction effects of the heartbeat (/), right bundle branch block (R), ventricular premature beats (V)

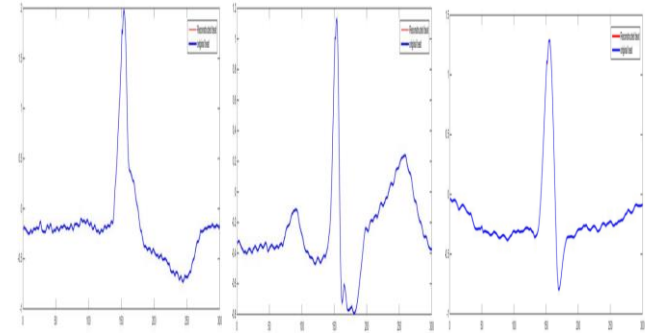

Figure 7. From left to right respectively are: the reconstruction effect of atrial premature beats (A), left bundle branch block (L), normal heart beat (N)

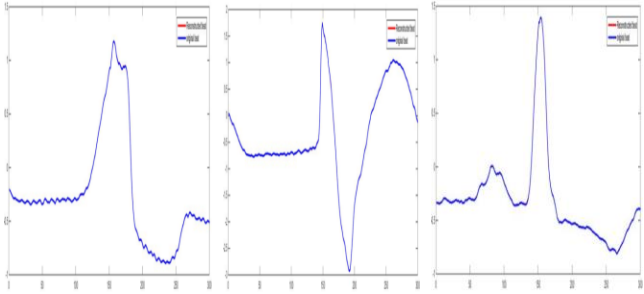

Figure 8. From left to right respectively are: the reconstruction effects of the heartbeat (/), right bundle branch block (R), ventricular premature beats (V).

TABLE II. COMPRESSION RATIO OF OUR SPARSE DICTIONARY ALGORITHM

| Heartbeats | N | / | A | V | R | L | Mean |
|---|---|---|---|---|---|---|---|
| Cr | 55.5% | 56.5% | 59% | 51% | 55.5% | 55.5% | 55.5% |

• Abnormal heart beats recognition

We use our method contrast with the traditional BOW heart beat model, and the traditional method of extracting feature points (ECG morphology and temporal features) and VQ method. All of them use SVM as a classifier. We compare the above six category heart beat recognition rates, as shown in Table III, and we calculate the accuracy of the two types:

All beats accuracy (the average accuracy for all beats) and Abnormal beats accuracy (the average accuracy for all abnormal beats), as shown in Figure 9.

TABLE III. HEART BEAT RECOGNITION RATE

| Method | N | / | A | V | R | L |
|---|---|---|---|---|---|---|
| Our method | 90% | 99.7% | 90% | 95.9% | 96.46% | 96.17% |
| BOW | 85% | 92% | 86% | 95% | 90% | 94% |
| Traditional method | 87.76% | 83.47% | 87.39% | 81.48% | 95.98% | 87.49% |
| VQ | 86.61% | 92.29% | 86.22% | 92.12% | 87.09% | 90.77% |

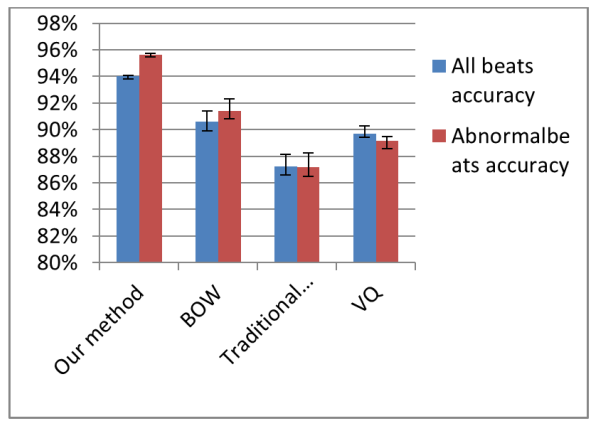

Figure 9. Two types of accuracies achieved on the test beats, where error bars represent standard deviations

We can see that our algorithm of heart beat recognition rate are achieved significantly improvement, especially for abnormal heart beats, the average accuracy are over 96%, and the pacing heartbeat (/) accuracy is nearly 100%. The all beats recognition accuracy is 94%, which is 3.4% higher than BOW method's, 6.74% higher than traditional method's, and 8.82% higher than VQ method's.

In summary, the proposed method also outperforms other methods in both accuracy and reconstruction effects.

## VI. CONCLUSION

In this paper, we propose the online sparse dictionary learning and Time Pyramid Matching algorithm, with the advantage of accuracy increasing and reconstruction error minimization in ECG classification system. We construct sparse dictionary, the heart beats are stored as the sparse coefficients, in this way not only meets the requirements of high-fidelity storage, but also reduce nearly 50% memory consumption. We propose histogram representation based on Time pyramid and stochastic pooling, our feature is more efficient, and achieve highest accuracy in ECG beats classification. All in all, due to above advantages, our approach has a high practical value and can be applied to long term monitoring and early warning for cardiovascular patients.

ACKNOWLEDGMENT

This work was supported by the Key Scientific and Technological Research Project of Jilin Province under Grant [number 20150204039GX and 20170414017GH]; the Natural Science Foundation of Guangdong Province under Grant [number 2016A030313658]; Innovation and Strengthening School Project (provincial key platform and major scientific research project) supported by Guangdong Government (2015KTSCX175); the Zhuhai Premier-Discipline Enhancement Scheme and the Guangdong Premier Key-Discipline Enhancement Scheme.